\def\year{2020}\relax
\documentclass[letterpaper]{article} 
\usepackage{aaai20}  
\usepackage{times}  
\usepackage{helvet} 
\usepackage{courier}  
\usepackage[hyphens]{url}  
\usepackage{graphicx} 
\usepackage{graphicx}  
\usepackage{multirow,bigdelim}
\usepackage{tabulary}

\title{Explainable Representations of the Social State:\\ A Model for Social Human-Robot Interactions.}

\author{Daniel Hernandez Garcia, Yanchao Yu, Weronika Siei{\'n}ska,\\ \Large \textbf{Jose L. Part, Nancie Gunson, Oliver Lemon, Christian Dondrup}\\
School of Mathematical and Computer Sciences\\
Heriot-Watt University\\
Edinburgh, Scotland, UK
}

\begin{document}

\maketitle

\begin{abstract}
In this paper, we propose a minimum set of concepts and signals needed to track the \textit{social state} during Human-Robot Interaction. 
We look into the problem of complex continuous interactions in a social context with multiple humans and robots, and discuss the creation of an explainable and tractable representation/model of their social interaction. 
We discuss these representations according to their representational and communicational properties, and organise them into four cognitive domains (scene-understanding, behaviour-profiling, mental-state, and dialogue-grounding).

\end{abstract}

\noindent Understanding the  world around us, and the intricate interactions that can take place in it, is a complex topic that has attracted interest from a large number of disciplines spanning philosophy, psychology, sociology, cognitive science, neuroscience, artificial intelligence, computer vision and robotics. 
While progress has been made (across all fields) and many theories, systems and architectures partially reproducing some parts of these cognitive skills have been developed \cite{Kotseruba201840YO},
the full mechanisms of cognition that explain these abilities in humans are still not completely understood \cite{FRITH2012}. 
Therefore, how to accurately obtain and interpret a representation of the social world remains a current problem for developing Socially Assistive Robots (SARs) \cite{defining_sar}. This problem is generally approached by the decomposition of the cognitive processes involved and a simplification of the interaction tasks.

\begin{figure}[t]
\centering
\includegraphics[width=0.725\columnwidth]{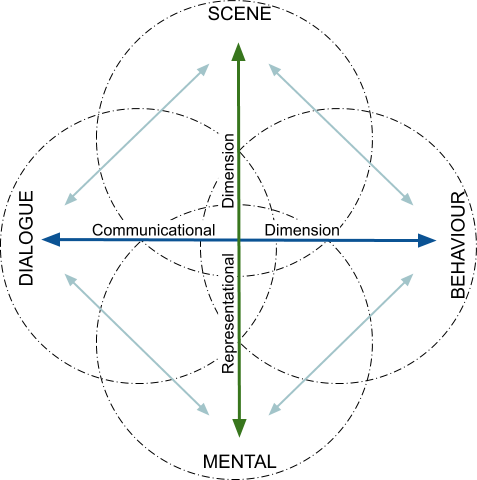} 
\caption{The space of the social-world model is grouped into 4 cognitive domains (scene-understanding, behaviour-profiling, mental-state and dialogue-grounding), organized at opposite ends across 2 functional dimensions: \textit{communicational}, verbal (dialogue) vs non-verbal (behaviour), and \textit{representational}, external (scene) vs internal (mental).}
\label{dimensions}
\end{figure}

In the last few years we can identify major advances in many of these topics, such as 
environment modelling \cite{dynamicSceneGraphs}, 
human activity recognition \cite{human_action_recognition}, 
speech recognition and speaker identification \cite{k2020joint}, conversational agents \cite{alana_v2}, natural language understanding \cite{VanzoBL19}, language grounding \cite{Yu_2017}, interactive task learning \cite{ijcai2018-1}, 
analysing interactions and behaviours \cite{TAPUS20193}, emotion recognition \cite{EGGER201935}, personality detection \cite{Mehta}, inferring intentions \cite{BiancoO19}, and
learning human-robot interactions \cite{Liu2018LearningPB}, 
among others. 
Based on the presented background, we define a minimal ``social state'' which enables continuous social interaction between SARs and humans by supporting reasoning and decision making during interaction with multiple agents and the environment. To this end, we {\it i)} group the above topics into areas of cognitive domains, presented along two dimensions (see Figure \ref{dimensions}) as we consider their representations and interconnections; and {\it ii)} present a minimal set of signals/information that will be needed to define a \textit{social state}. 
Maintaining such a model of the world and the social interactions is a complex task which makes decomposition and simplification a necessary abstraction. 
This abstraction, however, presents not just a challenge but an opportunity to not only reason about the state of the world and the agents within but also facilitate the creation of explainable decision-making engines that are not purely based on raw sensor data, as in most machine learning approaches, but on higher-level representations of the world.

\section{Model of the Social/World Interaction}

The main focus in Human-Robot Interaction (HRI) is often on the robot response, overlooking the way humans interact and preventing the robot from adapting to general situations \cite{Bianco2020}.
In order to address this, we need to be able model the whole interaction taking place in a tractable form that can inform other parts of the system. That is, the social nature of an agent should spread through its cognition by creating mechanisms for constructing social representations as an interpretation of the physical world, that allows processing of diverse social situations \cite{Rato2020TowardsSI}.

Developing SARs with the capacity of performing natural and continuous interactions in a social context with multiple agents in an `open-domain' requires the robot to show the ability to track and ascribe social meaning to its sensory information. 
SARs must explore the environment and understand what the environment affords, including  which objects, actions, events, and scene information can be extracted from the sensory data. They must track the state of each agent, and track their conversations, determining what they are saying and to whom, where their attention is at, as well as predicting their goals, and their affective and emotional states, etc. They must also `know' when to perform communicative actions and decide whom to respond to, and when and how to address one or more people. 
Developments in signal processing, behaviour analysis, multi-modal dialogue, machine learning, and robotics allow us to obtain rich sensory information from the world, but this is not enough as 
social signals are intrinsically ambiguous \cite{Vinciarelli2009SocialSP}, and agents must rely on the relationship between such elements \cite{Rato2020TowardsSI}. 
We suggest here, to broadly visualise these relationships according to the information that they contain by first categorizing then in terms of two dimensions.

\textbf{\textit{Communication dimension}}: human actions convey or express social information, either verbally, in dialogues and conversations, or non-verbally, in gestures, behaviours, pose, etc. We also know that the non-verbal behavior of an agent is critically important  as well as its verbal behaviour \cite{Vinciarelli2009SocialSP}. Hence, the social state must track both these domains to maintain conversation and interactions in their social context. 

\textbf{\textit{Representational dimension}}: social interactions are directed by what can be perceived or described from the environment as well as what can be predicted or estimated from the agent's internal belief, states, or desires. The social state must then keep both external (scene) and internal (mental) representations of the environment. 

We can see for each of these dimensions, the existence of two domains, differentiated by their nature as explained above. In this way we can organized the representations (see Figure \ref{dimensions}) into four domains (dashed circles) set across the two dimensions (solid lines). We propose this as an abstraction, as the dimensions are not to be considered as measurable spectrum but as a construct to distinguished the domains. So that the verbal (dialogue) and non-verbal (behaviour) domain can be intuitively separated on a \textit{communicational} dimension, and the external (scene) and internal (mental) domain can be separated on the \textit{representational} dimension. 
Furthermore, these domains are not excluding as they can be complementary to each other and can be used to help one another to extract meaning from/for the social context (interconnecting lines among domains), e.g. we can infer a person's intention by their gestures or conversation; dialogue can be grounded by what is known from the scene, or the person's goals; the estimation of a person's behaviour can be informed by the knowledge of the environment or the mental state attributed to them, etc.

Therefore, we propose the formation of four synergetic models for the: {\it 1)} representation of the scene; {\it 2)} representation of the person's behaviours;  {\it 3)} representation of the internal mental states; {\it 4)} representation of the conversations.

\subsection{Scene Understanding}

The semantic understanding of a scene is important for social robots applications. Spatial perception and 3D environment understanding are key enablers for high-level task execution in the real world. State-of-the-art approaches use the \textit{Scene Graph} paradigm, \cite{armeni_iccv19} provides a hierarchical 3D model that is useful for visualisation and knowledge organisation. 
\cite{dynamicSceneGraphs} propose an efficient scene representation data structure which can capture the environment from the lowest level represented as a metric-semantic mesh up through objects and agents in the environment up to rooms and buildings. 
Metric-semantic understanding provides the capability to simultaneously estimate the 3D geometry of a scene (critical for robots to navigate safely and to manipulate objects) and attach a semantic label to objects and structures (providing models of the environment for a robot to understand and execute human instructions) \cite{Rosinol20icra-Kimera}.

The model of the social scenario is a representation of the scene, see table \ref{table_scene}. Where is the interaction taking place? Who is taking part in the interaction? It models objects, places, structures, and agents and their relations in a way that is physically grounded from the environment by the robot's sensory information.

\begin{table}[ht]
\vspace{-0.5cm}
\caption{Representation of the scene.}\smallskip
\centering
\tymin=0.75in
\tymax=3in
\smallskip\begin{tabulary}{0.95\linewidth}{CCJ}
\bfseries Feature & \bfseries Signal & \bfseries Description \\
\hline
locale & metric-semantic, localisation & Where the interaction is taking place. Different rooms can require different interaction strategies. \\
agents & detection & List of agents in the scene (attended to or not).\\
objects & detection & List of (salient) objects. With attributes and affordances, etc. Tracked in the interaction. \\
rooms & metric-semantic & List of places (locales). \\ 
scene & scene-graph & Graph/map of the scene. Represents spatial concepts (objects, rooms, agents) and  spatio-temporal relations.\\
\end{tabulary}
\label{table_scene}
\vspace{-0.5cm}
\end{table}

\subsection{Behaviour Profile}

The ability to recognise and model physical human activities is a key technology to enable the development of useful HRI applications. The work of \cite{rossi_user_2017} provides key themes in the context of user profiling mechanisms and behavioral adaptation from the physical, cognitive and social interaction viewpoints. \cite{Aggarwal_Ryoo} provided a classification of the various types of human activities into four different levels: gestures, actions, interactions, and group activities. State-of-the-art solutions on computer vision and deep learning allow analysing the status of each person in the scene, based on body and head pose estimation, face recognition, facial landmarks extraction and the estimation of soft biometric patterns.

The model/profiling of the behaviour is a representation of the people interacting in the scene, see table \ref{table_behaviour}, and combines persistent data of the user for identification, i.e., name, id, role, and dynamic data of the user behaviour, i.e., current activity, focus of attention, location, status.  

\begin{table}[ht]
\vspace{-0.5cm}
\caption{Representation of the person's behaviours.}\smallskip
\centering
\tymin=0.75in
\tymax=3in
\smallskip\begin{tabulary}{0.95\linewidth}{CCJ}
\bfseries Feature & \bfseries Signal & \bfseries Description \\
\hline
ID & detection & ID of the person.  \\
group ID & detection & Group they belong to. \\
role & behaviour analysis & People could belong to different roles depending on interaction context. \\
activity & behaviour analysis & Track current activity.\\
attention & gaze & Track focus of attention.\\
location & localization & Track current location.\\
\end{tabulary}
\label{table_behaviour}
\end{table}

\subsection{Mental State}

Mentalising, mentalisation, or theory of mind refers to our ability to read the mental states of other agents \cite{FRITH2006531}. Findings in developmental psychology concerning current computational theories describing intention understanding and mental state inference from observed actions has inspired the development of architectures for social robots \cite{Bianco2020}. In \cite{BiancoO19}, a summary is provided on how theory of mind features have been integrated in robotic architectures for HRI.
\cite{pmlr-v80-rabinowitz18a} designed a neural network which uses meta-learning to build such models of the agents, able to predict the behavior of multiple agents in a false-belief situation given their past and current trajectories. 
Developing robots with mentalising capabilities for belief understanding, proactivity, active perception and learning of human behavior will further enhance robots’ capabilities and improve HRI \cite{BiancoO19}.

The model of the mental state is a representation of the internal model of the robot (agent) state, see table \ref{table_mental}.
It must track the intentions and beliefs of participants of the interaction, as well as predictions of the goals, motivations and emotional states, etc.

\begin{table}[ht]
\vspace{-0.5cm}
\caption{Representation of the agent's internal mental states.}\smallskip
\centering
\tymin=0.75in
\tymax=3in
\smallskip\begin{tabulary}{0.95\linewidth}{CCJ}
\bfseries Feature & \bfseries Signal & \bfseries Description \\
\hline
purpose & mentalising & Agent's main goal. \\
current target & mentalising & Goal pursued (by the agent) at present. \\
emotion state & behaviour analysis & Estimation of agent affective expression. \\
motivation & behaviour analysis & Condition of the agent, i.e, engaged, busy, waiting, etc.\\
\end{tabulary}
\label{table_mental}
\vspace{-0.5cm}
\end{table}

\subsection{Dialogue Grounding}

Humans use Natural Language (NL) to enable interpersonal communication, and articulate their thoughts and intentions. 
Social robots deployed in diverse human settings will need to interpret and execute high-level instructions given by NL.

Research in language grounding focuses on solving the symbol grounding problem for situated robots by leveraging their interactions with the humans they are working to understand \cite{thomason:jair20}. 
How can we talk to robots about the surrounding world? Can we enable them to interactively learn the grounded meanings needed to finish a task?  How can we assist a robot in their navigation or manipulation task with language instructions? Humans and robots need to bridge the gap in their representations to build a common ground of the shared world, for social robots to be able to engage in language communication and joint tasks \cite{Chai_Fang_Liu_She_2017}. 
Robots and humans will need to negotiate, using NL, the co-construction of shared representations and plans. This requires the creation of a unified and robust plan modelling and execution framework to combine dialogue actions and physical actions in the same planning domain of the human-robot social interaction \cite{dondrup2019petri}.

The model of the dialogue state is a representation of the conversation, see table \ref{table_dialogue}, tracking what has been said and by whom, the intents, the entities, the topics during interactions among multiple agents.

\begin{table}[ht]
\vspace{-0.5cm}
\caption{Representation of a conversation dialogue turn.}\smallskip
\centering
\tymin=0.75in
\tymax=3in
\smallskip\begin{tabulary}{0.95\linewidth}{CCJ}
\bfseries Feature & \bfseries Signal & \bfseries Description \\
\hline
speaker ID & detection & ID of the speaker. \\
listener IDs & detection & IDs of the listeners. \\
intent & NLU & The intent of the speaker. \\
entities & NLU & Entities in the dialogue. \\ 
topic & NLU & The current topic/task. \\
onset & ASR & When the dialogue starts. \\
transcript & ASR & Transcript of the dialogue.\\
\end{tabulary}
\label{table_dialogue}
\vspace{-0.5cm}
\end{table}

\subsection{Memory}
So far we have not think about how \textit{memory}, or \textit{time}, relates to the rest of the representations. 
One way to look at it is as a third dimension, orthogonal to the other two, so the \textit{social state} could be view as a slices of a plane running in a \textit{Memory} domain of current and past \textit{social states}. 
For now, however, we view temporal relations as ``integrated'' on the representations, and consider the representations on the \textit{social state} to correspond only to the present \textit{state} in time, and such we don't include an explicit model for \textit{Memory}.

\section{Discussion}

In order to create robots able to move, see, hear and communicate in a social context with multiple agents, and properly fulfil social roles and successfully execute social tasks, we need to model the human-robot interactions into an explainable and tractable representation of the social state.

Consider a social robot, tasked with interacting with patients at a clinic/hospital waiting room, in the scenario were a person comes in an approaches the robot. This immediately creates a near infinite number of decisions requiring person identification, tracking, intent beliefs, scene and behavior understanding, etc. It could be a new person, e.g. entering for the first time, and the robot would be required to welcome it, register their details and explain them a procedure to follow next. Or it could be a previously known person, e.g. coming back into the room from having been examined, asked to wait again for a follow up, here it would be unlikely that the robot is required to intervene but a social response acknowledging then back can be desirable. It could also be one of the persons waiting in the room, e.g. coming up to the robot to ask a question like \textit{"where is there a cafeteria or a restroom, or how long till they see them?"}.

Ideally a robot would be capable of providing different responses to all this scenarios. 
Deep learning approaches could be used to train an agent to learn the proper social interaction strategies \cite{Nanavati2020AutonomouslyLO}, \cite{romeo2019hai}. Also multiple deep neural networks can be used to generate the components that will provide inputs to this systems, such as scene understanding \cite{Zhang_2017_ICCV}, face detection \cite{abs-1902-03524} and natural language processing \cite{VanzoBL19}.

These are notorious for being black-box models that are hard, if not impossible, to interpret and which require explanations. Understanding this explanations will be facilitated by having then relate to the descriptions of the \textit{social state}.  
Successful human-robot social interactions will required not only that robots be able to create internal representations of the physical world, and of collaborative plans about that world, but also that they are able to communicate and negotiate about these representations in a manner that humans can understand.

For instance, in the above scenarios, the \textit{Scene} and \textit{Behaviour} domains would be fundamental to recognize new from known persons in the waiting room which determines different strategies to use to start an interaction with them.  The \textit{Mental} domain, to infer a person's interest and needs, is central to direct the best actions and goals the robot should pursue. The \textit{Conversation} domain is essential to lead the both user and robots in helpful dialogues. The synergies between these representations are crucial to generate fruitful interactions, e.g. when asked \textit{"where can I find a cafeteria?"} an advanced \textit{Conversation} representation is need to handle such queries, only relying on \textit{Scene} representation can the information required to answer be extracted, accurate \textit{Mental} representations can allow answer that can satisfied the needs of the user, e.g. the person wants a drink and the robot can answer \textit{"there's a vending machine at X"}, \textit{Behaviour} representation can indicate progress of the interaction, e.g. the person is following the indication from the robot, etc.

The ideas presented here, therefore, constitute a first step towards building a decision-making architecture for multi-party HRI and will be used as the basis for future work on SARs in healthcare.

\section{ Acknowledgments}

This work was funded by the EU H2020 programme -- 
grant agreement no. 871245 (SPRING).


\end{document}